\documentclass[a4paper,fleqn]{cas-dc}
\usepackage{xcolor}
\usepackage{enumitem}
\usepackage{amssymb}
\usepackage{flushend}
\usepackage{balance}
\usepackage{adjustbox}
\usepackage{float}
\usepackage{algpseudocode}

\usepackage[ruled,vlined,noresetcount ]{algorithm2e}
\usepackage{pifont}
\usepackage{tikz}
\newcommand*\circled[1]{\tikz[baseline=(char.base)]{\node[shape=rectangle,draw,inner sep=2pt] (char) {#1};}}

\usepackage{graphicx}
\usepackage{subfig}

\usepackage{multirow}

\def\tsc#1{\csdef{#1}{\textsc{\lowercase{#1}}\xspace}}
\tsc{WGM}
\tsc{QE}
\tsc{EP}
\tsc{PMS}
\tsc{BEC}
\tsc{DE}
\usepackage{float}
 \usepackage[usenames,dvipsnames]{pstricks}
\usepackage{pstricks-add}
 \usepackage{epsfig}
 \usepackage{pst-grad} 
 \usepackage{pst-plot} 
 \usepackage[off]{auto-pst-pdf}
 \usepackage[space]{grffile} 
 \usepackage{etoolbox} 
 \makeatletter 
 \patchcmd\Gread@eps{\@inputcheck#1 }{\@inputcheck"#1"\relax}{}{}
 \makeatother
\usepackage{auto-pst-pdf}
\usepackage{verbatim}
\usepackage{multicol}
\usepackage{epstopdf}
\usepackage[numbers]{natbib}

\begin{document}
\let\WriteBookmarks\relax
\def\floatpagepagefraction{1}
\def\textpagefraction{.001}

\shorttitle{Multilinear subspace learning for Person Re-Identification based fusion of high order tensor features}

\shortauthors{A. Chouchane ,et al.}

\title [mode = title]{Multilinear subspace learning for Person Re-Identification based fusion of high order tensor features}                      

\vskip2mm

\author[1]{Ammar Chouchane\corref{cor1}}
\ead{chouchane_ammar@yahoo.com}

\author[2]{Mohcene Bessaoudi}
\author[1]{Hamza Kheddar}
\author[2]{Abdelmalik Ouamane}
\author[3]{Tiago Vieira}
\author[4,5]{Mahmoud Hassaballah}

\address[1]{Department of Electrical Engineering, University of MEDEA, Algeria.}

\address[2]{Department of Electrical Engineering, University of Biskra, Algeria.}

\address[3]{Institute of Computing, Federal University of Alagoas, Brazil.}

\address[4]{Department of Computer Science, College of Computer Engineering and Sciences, Prince Sattam Bin Abdulaziz University, AlKharj, Saudi Arabia}

\address[5]{Department of Computer Science, Faculty of Computers and Information, South Valley University, Egypt.}

\tnotetext[1]{The first is the corresponding author.}

\begin{abstract}
Video surveillance image analysis and processing is a challenging field in computer vision, with one of the most difficult tasks being Person Re-Identification (PRe-ID). PRe-ID aims to identify target individuals who have already been recognized and appeared in a camera network, using 
a robust description of their pedestrian images. The success of recent research on PRe-ID is largely due to effective feature extraction and 
representation, along with powerful learning methods that enable reliable discrimination of pedestrian images. To this end, two powerful 
features—Convolutional Neural Networks (CNN) and Local Maximal Occurrence (LOMO)—are modeled on multidimensional data using the proposed method, 
High-Dimensional Feature Fusion (HDFF). Specifically, a new tensor fusion scheme is introduced to combine the two types of features in a single 
tensor, even though their dimensions are not identical. To improve our system's accuracy, we use Tensor Cross-View Quadratic Analysis (TXQDA) 
for multilinear subspace learning, followed by cosine similarity for matching. TXQDA efficiently facilitates learning while reducing the high 
dimensionality resulting from high-order tensor data. The effectiveness of our approach is verified through experiments on three widely used PRe-ID 
datasets: VIPeR, GRID, and PRID450S. Extensive experiments demonstrate that our approach performs very well compared to recent state-of-the-art methods.
\end{abstract}




\maketitle

\section{Introduction}
\label{sec:introduction}

\begin{figure*}
	\centering
	\includegraphics[width=.9\textwidth]{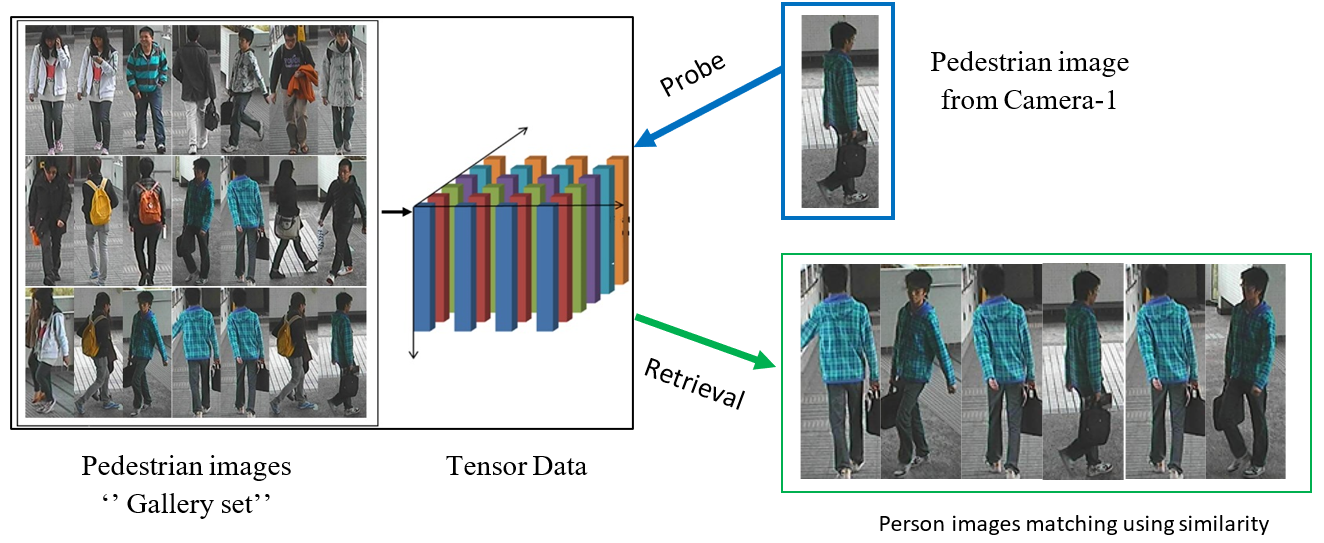}
	\caption{Person Re-Identification pipeline.}
	\label{fig:pipeline}
\end{figure*}

Recently, the scientific community of computer vision and artificial intelligence has become increasingly interested in Person Re-Identification (PRe-ID) due to its various security-based applications, beyond biometrics techniques \cite{hemis2024deep},  such as: Cross-camera person tracking by detection; person retrieval; human-machine interaction Long-term human behavior and activity analysis \cite{Suma2019COMPUTERVF,Perwaiz2018PersonRU,hassaballah2019}. Besides, the video surveillance using different cameras installed in public places is mainly used for monitoring the behavior and identification of suspects in public transport like railway stations, airports, inside planes and trains etc \cite{Bolle2004,hassaballah2020deep}, as well as in international events such as  World Cup, Olympic Games, etc. 
PRe-ID refers to the task of matching pedestrians observed from non-overlapping views of surveillance camera networks~\cite{sharma2015}. This system has widely studied during the recent years, however, it still a difficult and a challenging task. It is habitually done by extracting discriminating features from low-resolution pedestrian images. However, the difficulty level of this task is increased because the captured views using the surveillance cameras are recorded under non-controlled conditions, and thus, the obtained images may contain large variations due to changes in poses, low resolution, illuminations, camera viewpoint variations, similar clothing, expression variations and partial occlusions~\cite{sharma2015,7298832,7849147,8953522}. 

As shown in Figure 1, the gallery set contains multiple pedestrian images of a target person captured by other cameras in addition to other persons. The query person image is treated and matched with the gallery set images in order to determine whether the person's images are present in the gallery set.PRe-ID system accomplishes three crucial tasks in video surveillance, including : person detection, person tracking, and person retrieval~\cite{wu2019deep}, which emphasizes importance to improve this systems by the research community. In this paper, we are interesting by person retrieval task in which the data is represented with the high odrer tensors.  All the details about the processing of this data will be presented in following sections.

$k$-fold cross-validation is the protocol used with the datasets of PRe-ID, the learning pedestrian set is partitioned randomly into $k$ disjoint subsets. Then, the model is trained using $k$-1 folds, which represent the training set. Then, the model is applied to the remaining folds, which is denoted as the validation set, in which, the performances is calculated. This procedure is repeated until each of the $k$-fold has served as validation set. The average of the $k$ performances measurements on the $k$ validation sets is the cross-validated performance. Cumulative Matching Characteristics (CMC) is the graphical representation of the identification results, plotting Rank value on the $x$-axis, and the probability of correct identification on the y-axis. PRe-ID focuses on matching the same person across non-overlapping camera views.

This paper specifically suggests an attention-aware multidimensional feature fusion based multilinear subspace learning approach for the PRe-ID task. We believe that Local Maximal Occurrence (LOMO) descriptor with its robustness description to colors and texture invariant to the challenge of illumination variations; additional to Convolutional Neural Network (CNN) with its powerful non-linearity will provide a significant discrimination power resulting in extremely satisfying results. For multilinear subspace learning, we used a new method called Tensor Cross-View Quadratic Analysis (TXQDA), which has been proven to be very effective with kinship verification systems~\cite{LAIADI2020286}. 

The main contributions of this paper are:


\begin{itemize}
	\item A new features model that fuses CNN and LOMO descriptors is proposed using multidimensional data representation considering high order tensors.
	
	\item Tensor Cross-View Quadratic analysis is utilized to perform powerful multilinear subspace learning followed by the Cosine similarity for matching. TXQDA efficiently ensures the learning and reduces the high dimensionality resulting from high order tensors, this, leads to improve the discrimination power. 
	
	\item The obtained results on three challenging datasets, VIPeR, GRID and PRID450S are compared against the recent state-of-the-art methods.
	
\end{itemize}
The rest of the paper is organized as follows. Section \ref{sec:LS} gives a literature survey about shallow feature extraction methods and the recent CNN deep features in PRe-ID in addition to an overview about multidimensional metric and subspace learning. Section \ref{sec:High-DFF} presents the proposed High-Dimensional Feature Fusion scheme. Section \ref{sec:TXQDA} presents a thorough illustration of the multilinear subspace learning algorithm TXQDA. Experiments and results analysis are presented in section \ref{sec:results}. Finally, Section \ref{sec:conclusion} draws our conclusions.

\section{Literature survay}
\label{sec:LS}

Fundamental problems in any PRe-ID are feature extraction and the learning.  In this section, a recent researches about PRe-ID system are reviewed with a particular focusing on the most important shallow features and CNN based deep features. Furthermore, to enrich the paper, multilinear subspace learning algorithms in several biometric applications are discussed.

The two efficient and famous methods of feature extraction for PRe-ID, are Local Maximal Occurrence Feature (LOMO) ~\cite{7298832} and Gaussian-of-Gaussians(GOG) ~\cite{7780521}. Considering non-uniformities inherent to PRe-ID systems due to different camera hardware and image acquisition settings, Liao et al~\cite{7298832} proposed a hand-crafted attribute coined LOMO to alleviate such variations. Indeed, perceived colors of a given individual across different frames and camera views present large variations between pedestrian images, which makes it difficult to identify them. By applying a preprocessing step based on a color normalization algorithm capable of providing a color image more consistent to human perception, HSV histograms can be properly used as discriminative features. They combined color descriptors with Scale Invariant Local Ternary Pattern (SILTP) ~\cite{5539817} to incorporate texture description invariant to illumination variations. In addition, to tackle changes in viewpoints, SILTP features are computed on sub-windows for local feature extraction and better characterization of a pedestrian images. The final goal is to achieve color, viewpoint changes and scale invariance by concatenating HSV and SILTP features computed on a pyramid representation of an image. After feature normalization via log transform, the final vector has 26960 dimensions and is not expensive to compute. Using LOMO feature descriptor, experiments on VIPeR dataset, provide a performance improvement of 2.2\% percentage points compared to previous approaches. 

Bazzani et al.~\cite{5539926} focused on human appearances conveyed by body cues into three complimentary characteristics; overall, spatial distributions of colors, and the presence of recurrent high entropy motifs. They proposed a feature descriptor composed by Symmetry-Driven Accumulation of Local Features (SDALF), which presented a good performance in different PRe-ID datasets. To determine whether two person-centered Bounding Box correspond to the same person, Ma et al.~\cite{10.1007/978-3-642-33863-2_41} proposed an attribute descriptor colled Local Descriptors encoded by Fisher Vector (LDFV). The latter describes each pixel as a 7-d local feature, comprised by its coordinates, intensity, first and second-order derivatives around its neighborhood. LDFV achieved a good  robustness to changes in illumination, viewpoint, background noise, occlusions and low resolution.
Matsukawa et al~\cite{7780521} noticed that information conveyed by the average value of pixels belonging to sub-windows are important for the discrimination of pedestrian images. Also, said statistic is not considered when covariance feature descriptors are used to represent the structures of these images. To achieve more efficiency on feature extraction, they proposed a descriptor coined Gaussian-of-Gaussians (GOG) which has the ability to describe global characteristics using the local distribution of pixel features. Indeed, people’s clothes usually consist of a few predominant colors distributed throughout garments, indicating that local average pixels values can provide a good discriminative capacity for pedestrian image discrimination. 

Furthermore, recent studies are directed towards the use of CNN deep features to improve PRe-ID 
~\cite{7780521,xie2022sampling,10.1016/j.patcog.2021.108287, yao2019deep, zheng2017discriminatively, liu2017end, xuan2022intra, khan2021deep, jayavarthini2022deep}. Matsukawa et al ~\cite{7780521}, resolved the problem of person PRe-ID based on learning of CNN features extracted from of pre-trained ImageNet dataset. fine-tuning attributes is used on pedestrian images in order to improve CNN features and give more discriminate power. Yao et al. ~\cite{yao2019deep} proposed a new deep representation learning process called part loss network to get more discrimination to the pedestrian images. This approach has achieved two advantages by minimizing two risks: the empirical classification risk on training person images and the representation learning risk. Always in the direction of constructing and improving discriminative learned features, Xuan et al. ~\cite{xuan2022intra} address an unsupervised person re-Identification system based on decomposing the similarity computation into two stages, intra-domain and inter-domain. The learned CNN features within each camera, are leveraged by intra-domain similarity, and a final feature vector is constructed by the inter-domain similarity scores of each sample on different cameras. Khan et al. ~\cite{khan2021deep} propose to improve person re-identification system using a new framework-based CNN deep features using self-tuned autoencoder. Indeed, the authors subdivise the input image into two patches, upper and lower in which the learned features are obtained from the fully connected layer of the CNN model. On the other hand, to obtain high performance, a self-tuned autoencoder is used to get low-dimensional features.

Jayavarthini et al. ~\cite{jayavarthini2022deep} introduce a new architecture of deep learning named  Expanded Neighbourhood Distance Reranking (ENDR) based on Densely CNN (DenseNet169) to improve the accuracy of PRe-ID against the appearance perturbations. Raj et al.~\cite{10.1016/j.patcog.2021.108287} suggested an unsupervised approach for PRe-ID called, Spatio-Temporal Association Rule based Deep Annotation-free Clustering (STAR-DAC). The unlabeled pedestrian images are labled by STAR using fine-tune algorithm.  To solve the visual similarity problem, STAR-DAC uses two principal process, outlier elimination and reliable sample selection. This method has achieved excellent results with large scale datasets.

Discriminant metric learning process is a very important phase in PRe-ID. In this regard, multilinear subspace learning using high order tensor data give great strength to many biometric applications such as kinship verification, gait recognition, and 2D and 3D face recognition. During the processing of the image data, images must be converted to a vector and stored as a matrix, in which, the spatial correlations are broken. On the other hand, in the most cases the number of samples is rather small related to the feature vector dimension ~\cite{https://doi.org/10.48550/arxiv.2205.09191}. In order to deal with these problems, many multilinear subspace projection and dimensionality reduction have been proposed to use in different images processing applications. Lu H et al. ~\cite {4359192} extended the famous Principal Component Analysis(PCA) to a multilinear version named Multilinear Principal Component Analysis (MPCA) to deal with tensor data. MPCA is applied to the problem of gait recognition, in which, good results are given. After that, Ouamane et al.~\cite {7954760} extend MPCA to Multilinear Whitened Principal Component Analysis (MWPCA), in which its covariance matrix become identity matrix where the data becomes less correlated. The same authors, in ~\cite {7954760} proposed another new version of Linear Discriminant Analysis (LDA) in order to address Small Sample Size (SSS) problem, named Tensor Exponential Discriminant Analysis (TEDA) using the notion of Matrix Exponential approach ~\cite {5184935} with the global projection matrices. MWPCA and TEDA are applied to 3D and 2D face recognition. 

Yan et al. ~\cite {4032832} extended LDA ~\cite {belhumeur1997eigenfaces} to Multilinear Discriminant Analysis (MDA) in order to improve discriminate power, MDA uses an iterative process to learn manifold interdependent discriminative subspaces as well as to reduce the dimensions of high order tensors. MDA is applied to 2D face recognition. 

Bessaoudi et al. ~\cite { BESSAOUDI2019267} proposed a new Multilinear Side-Information based Discriminant Analysis (MSIDA) for dimensionality reduction and classification of tensor data to improve the performance of kinship verification based on face images. The authors resolve the problem of face pair matching based on multidimensional tensor representation in the presence of weakly labeled training data. Very satisfactory results were obtained compared to the basic linear approach Side-Information based Linear Discriminant Analysis (SILD) ~\cite { Kan2011SideInformationBL}. In the same context of kinship verification, Laiadi et al. ~\cite {LAIADI2020286} proposed an efficient method for multilinear subspace learning called Tensor Cross-View Quadratic Analysis (TXQDA). The kinship verification problem is tackled as a cross view matching, in which TXQDA based tensor face data enlarges the margin between samples and helps also helps in solving a problem of SSS.

\begin{figure*}[!htb]
	\centering
	\includegraphics[height=4in, width=7in]{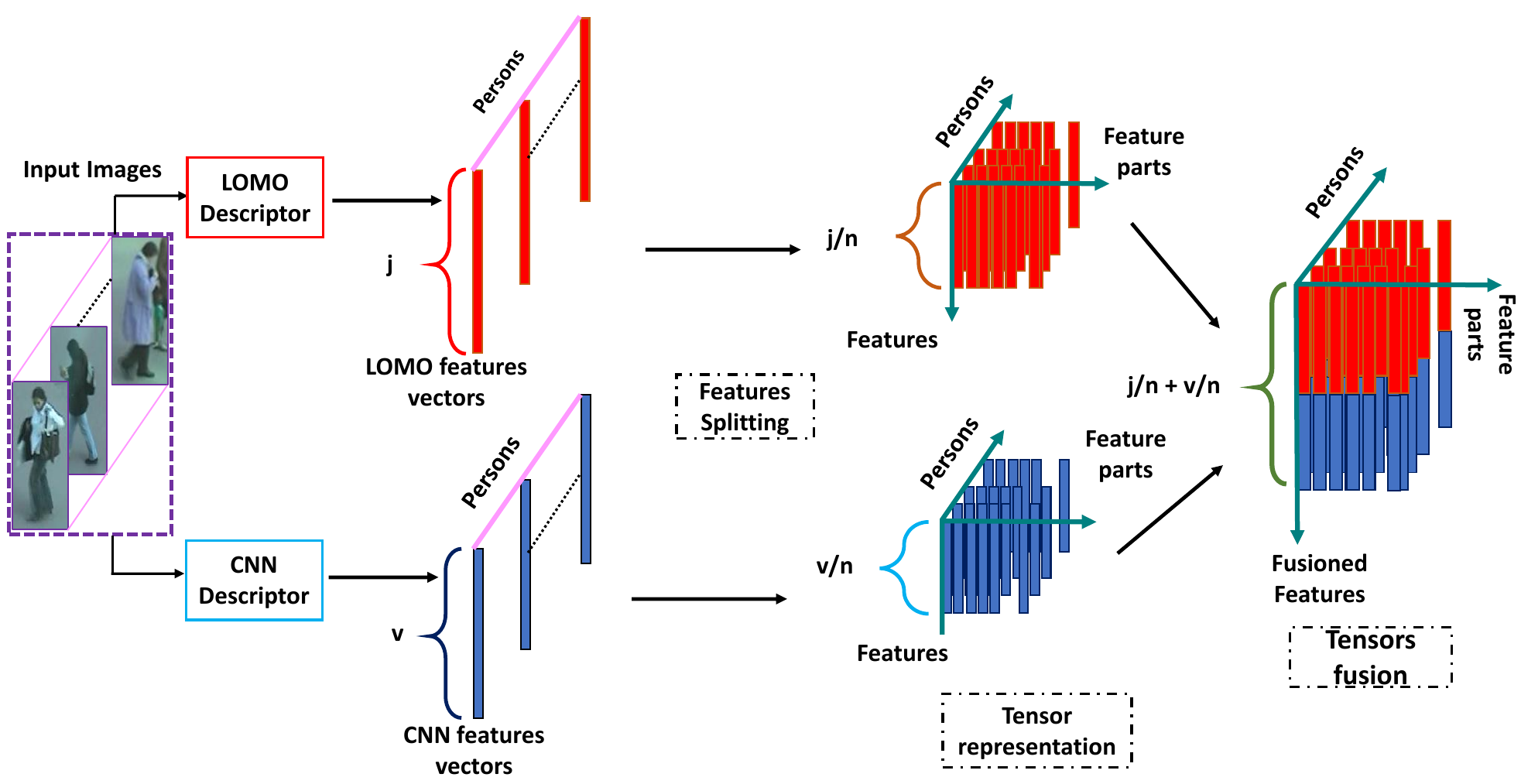}
	\caption{High-Dimensional Feature Fusion framework.}
	\label{HDFF1}
\end{figure*}


\section{Proposed High-Dimensional Feature Fusion}
\label{sec:High-DFF}

To make it easier to grasp the mathematical concept, this section highlights the mathematical notations used to deal with tensors of high order. After that, the proposed multidimensional fusion strategy is described. 

\subsection {Mathematical notations}
\begin{figure}[!htb]
\centering
\includegraphics[height=2.8in,width=3in]{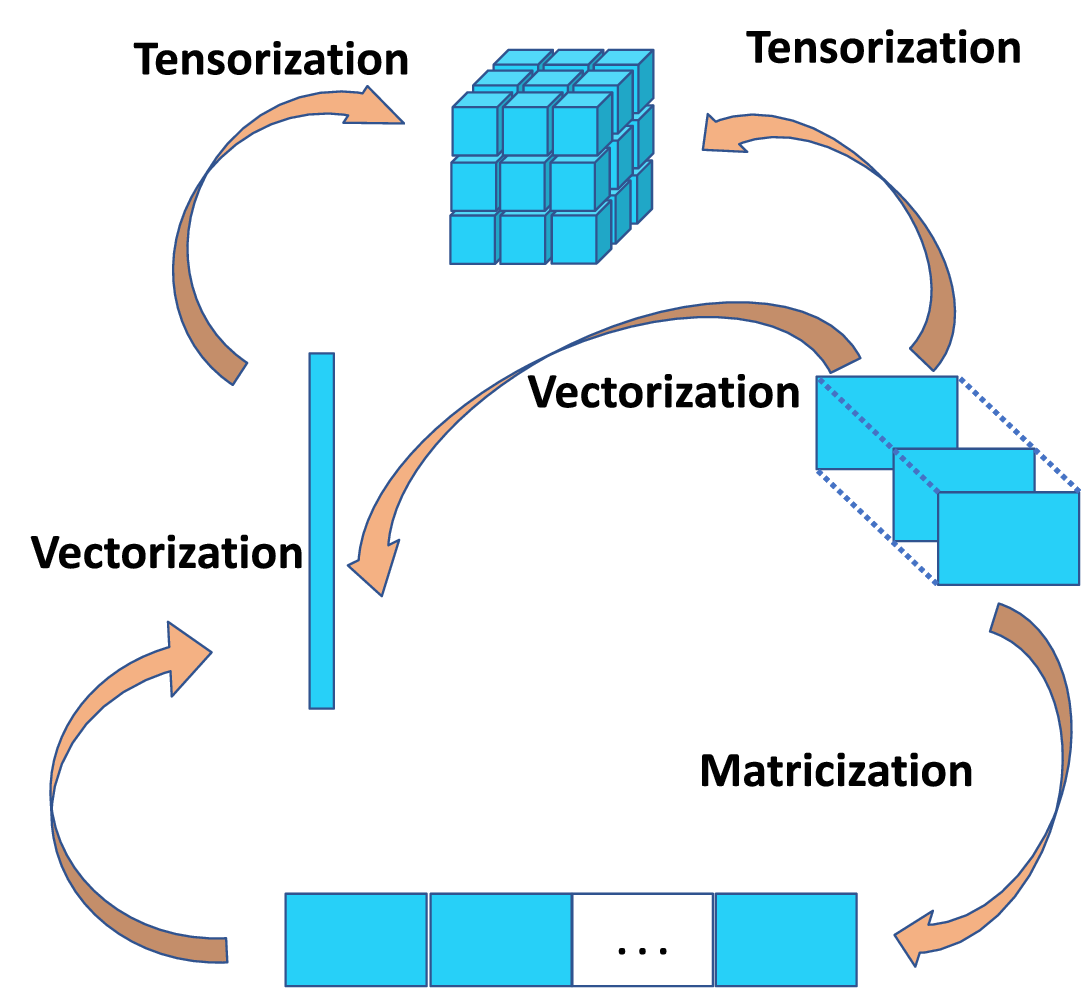}
	\caption{Tensor reshaping process: vectorization, Matricization and tensorization.}
	\label{fig:unfolding55}
\end{figure}

The variables and mathematical notations used in the proposed High-Dimensional Feature Fusion (HDFF) and TXQDA algorithm are defined as follows:

\begin{itemize}
	\item  Lowercase and uppercase symbols (e.g., i, j, K, L and $\lambda$ ) denote scalars. 
	\item  Bold lowercase symbols (e.g., \textbf{$x$}, \textbf{$y$} and \textbf{$\alpha$}) denote vectors. 
	\item  Italic uppercase symbols (e.g., \emph{U}, \emph{N}, \emph{X} and \emph{V}) denote matrices. 
	\item  Bold italic uppercase symbols (e.g., $\textbf{\emph{A}}$, $\textbf{\emph{B}}$, $\textbf{\emph{C}}$ and $\textbf{\emph{X}}$) denote the high order tensors.
\end{itemize}
The most frequently  multidimensional-related Operations used in this paper are listed in Table~\ref{talts}. Furthermore, figure~\ref{fig:unfolding55} shows three tensor reshaping process: vectorization, matricization (Unfolding) and tensorization. Vectorization refers to transforming a matrix or tensor into a 1D vector, matricization refers to transforming a tensor into a 2D matrix, whereas tensorization refers to transforming low-order tensors (vector or matrix) into n order tensor.

\begin{table*}[!ht]
	\centering
	\caption{Basic multidimensional operations}
	\label{talts}
	
	\begin{tabular}{|l|l|}
		\hline
		{\bf Tensor operations } & {\bf Algebratic formulas} \\
		\hline
		Mode-k tensor matricization &         $\mathrm{\textit{U}_k}\mathrm{\in\mathbb{R}^{{I_k}\times ({I_1}\times...\times I_{k-1}\times I_{k+1} \times...\times {I_k})}} $ \\
		\hline
		Mode-k tensor product &          $\textbf{\emph{C}}=\textbf{\emph{A}}\mathrm{\times{_k}\textit{U}}\mathrm{\in\mathbb{R}^{{m_1}\times {m_2}\times ...\times {m_{k-1}} \times {p_k} \times {p_{k+1}}\times {m_{k+1}}\times ...\times {m_n}}}$ \\
		\hline
		Tensor-to-tensor projection &         $\textbf{\textit{X}}\mathrm{=}\textbf{\textit{Y}}{\mathrm{\times }}_{\mathrm{1}}U^{(1)^\mathrm{T}}{\mathrm{\times }}_{\mathrm{2}}U^{(2)^\mathrm{T}}\times ... {\mathrm{\times }}_{\mathrm{n}}U^{(n)^\mathrm{T }}$ \\
		\hline
		Tensor vectorization   &          $x=\mathrm{vec}(\textbf{\textit{X}})$ \\
		\hline
		Tensor frobenius norm  &          $ ||\textbf{\textit{X}}||_\mathrm{F}=\sqrt{\left \langle \textbf{\textit{X}},\textbf{\textit{X}}\right\rangle} $  \\
		\hline
		Scalar product  &          $\left \langle \textbf{\textit{A}},\textbf{\textit{B}}\right\rangle =\mathrm{\sum_{i_{1},...,i_{N}}a_{{i_1}...{i_N}}.b_{{i_1}...{i_N}}}$ \\
		\hline
	\end{tabular}
	
\end{table*}







\subsection {Methodology}
\begin{figure}[ht!]
	\color{red}
	\begin{center}
		\includegraphics[scale=0.6]{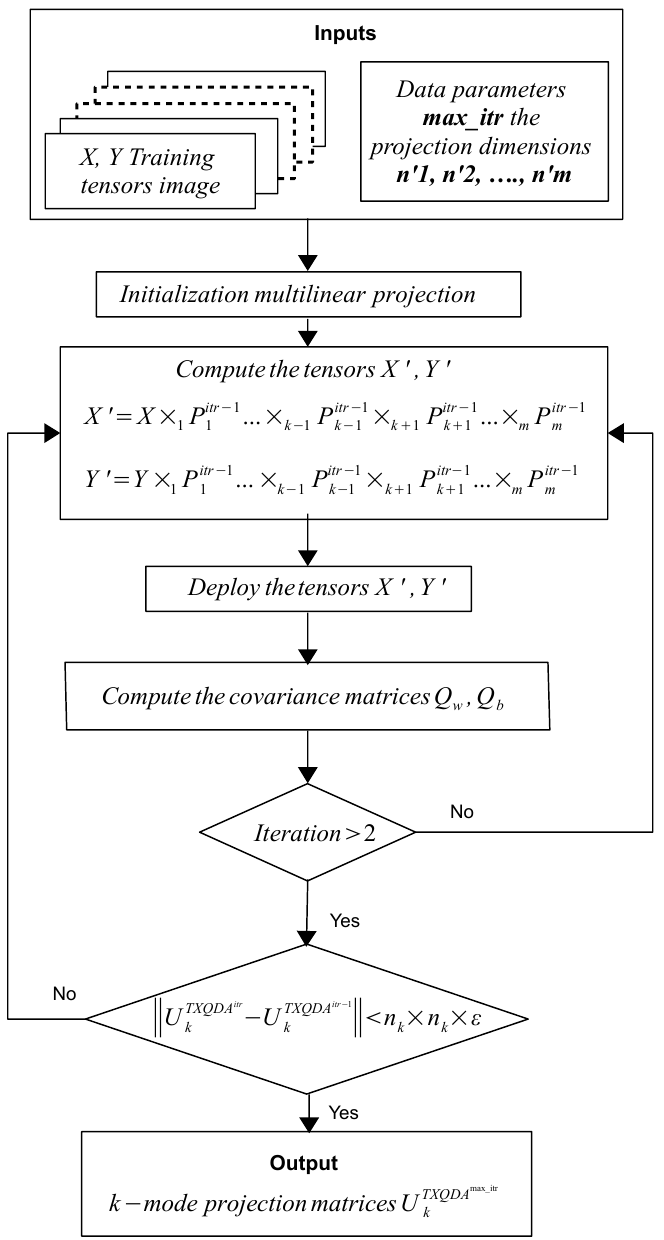}
	\end{center}
	\caption{The principle of TXQDA algorithm.}
	\label{figScheme}
\end{figure}

Exploit different heterogeneous features in biometric Recognition systems is a challenging task. Among the excellent solutions for this process is the multidimensional data representation based on high order tensors ~\cite{LAIADI2020286,7954760,4032832,bessaoudi2021multilinear}. Indeed, the dimensions in heterogeneous feature types are not the same, present a challenge to build suitable data for next processing steps. In this work, we propose a novel high-dimensional data construction approach in order to explore high-order correlations, such that one can take  advantage of extracted heterogeneous features. The schematic of tensor fusion strategy is shown in Figure ~\ref{figScheme}. 

Given two heterogeneous sequences of LOMO and CNN feature vectors $\mathrm{\textbf{x}=\{{x_1},...,{x_j}\}} \in\mathbb{R}^{j} $ and $\mathrm{\textbf{y}=\{{y_1},...,{y_j}\}} \in\mathbb{R}^{v}$. We split the features with same number of parts to obtain a set of n vectors $\mathrm{\emph{A}=\{{\textbf{a}_1},...,{\textbf{a}_n}\}\in\mathbb{R}^{{(j/n)}\times n}}$, $\mathrm{\emph{B}=\{{\textbf{b}_1},...,{\textbf{b}_n}}\}\in\mathbb{R}^{{(v/n)}\times n}$, the next step is to combine each kind independently as columns of a matrix $\emph{A}_\mathrm i \mathrm{\in\mathbb{R}^{(j/n)\times n}}$ and $\emph{B}_\mathrm i \mathrm{\in\mathbb{R}^{(v/n)\times n}}$ creating a second order tensor representation, in which, the n number parts represents the dimension of second mode for each kind of feature. The $\mathrm{n^{th}}$ column of $\emph{A}_\mathrm{i}$ and $\emph{B}_\mathrm{i}$ only consists of the $\mathrm{n^{th}}$  feature parts view from the $\mathrm{i^{th}}$  sample. By organizing data in this way, the set of  ${\{\emph{A}_\mathrm i}\}\mathrm{^{m}_{i=1}}$, ${\{\emph{B}_\mathrm i}\}\mathrm{^{m}_{i=1}}$ will lead to a combination of different views while regarding each person view data, where m is the totally views. Through using arranging manipulation, all individual view data is transformed into two $\mathrm{3^{rd}}$ order tensors representation space, that is, $\textbf{\emph{A}}\mathrm{\in\mathbb{R}^{(j/n)\times n \times m}}$ and $\textbf{\emph{B}}\mathrm{\in\mathbb{R}^{(v/n)\times n \times m}}$. After getting the two heterogenous tensors $\textbf{\textit{A}}$ and $\textbf{\textit{B}}$, unlike classic methods that rely on the fusion at the vector level, the proposed HDFF try to take advantage the inter-modal between the two tensors data through the fusion process at the tensor level along the first mode tensors (mode-1)  in which mode-1 represents the features vectors parts. the fusion strategy is modeled as: $\textbf{\textit{C}}$=HDFF ($\textbf{\textit{A}}$,$\textbf{\textit{B}}$), where $\textbf{\emph{C}}\mathrm{\in\mathbb{R}^{ s \times n \times m}}$ is the obtained $\mathrm{3^{rd}}$ order enhanced features tensor such that $\mathrm{s=\frac{j}{n}+\frac{v}{n}}$. and HDFF (.) is the proposed tensor fusion process. As a consequence, the high correlation can be perfectly exploited by using all views simultaneously in each tensor representation, as well as, the data structure in the obtained tensors provides more discriminative information. The proposed model for high-dimensional feature representation is summarized in Algorithm \ref{alg1}.


\begin{algorithm*}[ht]
	\textbf{Input:} Low level feature vectors $\textbf{x}_\mathrm{i}=\mathrm{\{x_1,\dots,x_j }\}\in \mathbb{R}^{j},\  \mathrm{i=1,...,m} $ \Comment{For LOMO descriptor}\\
	\hspace{1cm} Deep feature vectors, $\textbf{y}_\mathrm{i}=\mathrm{\{y_1,\dots,y_v}\}\in \mathbb{R}^{v},\ \mathrm{i=1,...,m}  $ \Comment{For CNN descriptor}\\
	\textbf{Output:} $\mathrm{3^{rd}}$ order tensor  $\textit{\textbf{C}}\in \mathbb{R}^\mathrm{{s\times n\times m} \Big( s=\frac{j}{n}+\frac{v}{n} \Big)}$\\
	\textbf{Steps:} 
	\begin{itemize}
		\item [1-] Break down LOMO feature vector \textbf{x} into $\mathrm{n}$ parts. 
		\item [2-] Break down CNN deep feature vector \textbf{y} into $\mathrm{n}$ parts.
		\item [3-] Combine the features vector parts to create \textit{\textbf{A}} and \textit{\textbf{B}} second order tensors.
		\item [4-] Arrange all view data to create \textit{\textbf{A}}$\in \mathbb{R}^\mathrm{{\frac{j}{n} \times n\times m}}$  and \textit{\textbf{B}}$\in\mathbb{R}^\mathrm{{\frac{v}{n}\times n \times m}}$ $\mathrm{3^{rd}}$ order tensors
		\item [5-] Fusion of \textit{\textbf{A}}$\in \mathbb{R}^\mathrm{{\frac{j}{n}\times n\times m}}$ and \textit{\textbf{B}}$\in\mathbb{R}^\mathrm{{\frac{v}{n}\times n \times m}}$  using concatenation procedure to obtain
		\item [6-] The enhanced tensor \textit{\textbf{C}}$\in \mathbb{R}^{\mathrm{s\times n\times m}}$ such that $\mathrm{s=\frac{j}{n}+\frac{v}{n}}$.
	\end{itemize}
	\caption{High-Dimensional Feature Fusion }
	\label{alg1}
\end{algorithm*}



\section{Multilinear subspace learning and matching}
\label{sec:TXQDA}
The fusion of tensors data based on different types of features resulting the problem of high-dimensional as well as enormous redundant information, which will affect the performance of discrimination in the classification or matching step. To resolve this issue (topic), we employ TXQDA method. This algorithm is an efficient multilinear weakly supervised framework used to reduce the dimensionality of high order tensors data and to enhance the discrimination power in addition to minimize the execution time. As a result, more feature dimensions are reachable in TXQDA than Cross-View Quadratic Analysis (XQDA) ~\cite{7298832}, because the feature dimension of XQDA is theoretically restricted by the number of classes, while with TXQDA, this problem is resolved. Moreover, among the strengths of this algorithm, the computational cost is drastically decreased because each step mode optimization is done on a smaller-sized feature space. TXQDA determines a set of projection matrices $U_\mathrm{k}$ that map the high order tensor $\textbf{\emph{X}}\mathrm{\in\mathbb{R}^{ s \times n \times m}}$ into new reduced tensor space $\textbf{\emph{X}'}\mathrm{\in\mathbb{R}^{ s' \times n' \times m'}}$ the target function of TXQDA is to decrease discriminability among samples from same classes via minimizing the covariance matrix $\emph{V}_\mathrm{I}^\mathrm{k}$  whereas maximize discriminability among samples from distinct classes via maximizing the covariance matrix $\emph{V}_\mathrm{E}^\mathrm{k}$ in multilinear formulation, as shown in the following equation, 

\begin{equation} \label{eq_11}
\mathrm{\textit{U}^*_k}=\mathrm{argmax}_{\textit{U}_\mathrm{K}}\frac{\mathrm{Tr}(\textit{U}^\mathrm{T}_\mathrm{k}V^\mathrm{k}_\mathrm{E}\textit{U}_\mathrm{k})}{\mathrm{Tr}(\textit{U}^\mathrm{T}_\mathrm{k}V^\mathrm{k}_\mathrm{I}\textit{U}_\mathrm{k})}
\end{equation} 

Fig.~\ref{figScheme} describes the TXQDA algorithmic procedure. There is no closed-form solution for the Discriminant Tensor Criterion (DTC). The difficulties are addressed by an iterative process in which the algorithm is based on the basic mathematical tools illustrated in table ~\ref{alg1}.

The target dataset is then subjected to multi-linear metric learning based TXQDA to improve discriminative power for PRe-ID. We used the Cosine similarity (CS). It is a prominent and widely used approach for improving matching results in several biometric recognition applications.Using cosine similarity in any given metric results in a powerful learning technique that can enhance the generalization capacity ~\cite{ nguyen2010cosine}. 

Given two vectors $\textbf{x},\textbf{y} \in \mathcal{R}^{m}$, in which m is the size of the two vectors, the goal is to estimate the CS distance such that:

\begin{equation}
\centering
\mathrm{CS} \left( \textbf{x},\textbf{y} \right) = \frac{\textbf{x}^{T} \textbf{y}}{||\textbf{x}||.||\textbf{y}||}
\label{eq:matrix-m}
\end{equation}

The 10-fold cross-validation strategy is used in the experiments; datasets are randomly divided into two subsets, training and testing. The test set is divided into the gallery and probe sets, in which the gallery set contains the images from camera one, and the probe set contains images from camera two. This procedure is repeated ten times. The average accuracy of the 10 measures is reported as an accuracy. The goal of PRe-ID is to identify the correct matching image of a probe in the gallery set using the trained model. The ideal case is that the correct matching image ranks first in the ranking list, i.e., the right matching image is the most similar to the probe image.


\section{Experiments and Results Analysis}
\label{sec:results}

To evaluate the effectiveness of the proposed approach, various experimental studies are performed, including comparisons with state-of-the-art methods, on three challenging PRe-ID datasets: VIPeR ~\cite{10.1007/978-3-540-88682-2_21}, GRID ~\cite{Loy2010} and PRID450S ~\cite{Roth2014}. The three datasets are briefly introduced in the following subsection.

\subsection {Datasets description}
We evaluated the performance of the CNN, LOMO and CNN+LOMO features using three challenging  datasets, VIPeR, GRID, and PRID450S, described in the following.



\begin{itemize}
	\item PRID450S dataset is collected using two  overlapping cameras. It contains 900 pedestrian images of 450 person, including pairs of walking pedestrians, recorded from two disjointed surveillance cameras. The  images size is $128 \times 48$ pixels. The challenges included in this dataset are: viewpoints, illumination variations, pose variations, and significant differences in background of a pedestrian images. 255 identities are randomly selected to build the training set, and the remaining 225 identities are used for the testing process \cite{roth2014mahalanobis}.
	
	\item VIPeR dataset is captured from two disjointed surveillance cameras. It contains 1264 pedestrian images of 632 person. Each  pair images was captured from a random viewpoint under different illumination variations. All images are resized to 128$\times$48 pixels. 316 identities are randomly selected to build the training set, and the remaining 316 identities are used for the testing process. This is one of the most challenging datasets for automated person PRe-ID \cite{bedagkar2014survey,leng2019survey}.
	
	\item GRID dataset was captured using 8 camera views. It contains 1275 images of 250 person. 775 images do not belong to the 250 people, which can be used to enlarge the gallery set. 125  pair images are used for training, and the remaining 125 pair images are used for testing. According to \cite{leng2019survey}, the size of the image range from 29 $\times$ 67 to 181 $\times$ 384. The low resolution strongly present in this dataset  which makes it one of a very hard datasets in Person Re-Identification system. 
	
\end{itemize}

\subsection{Evaluation metrics}
The CMC curve is the most recognized and preferred technique used for assessing the performance of PRe-ID algorithms in closed-set identification scenarios. Compared to the unknown probing image, CMC evaluates how effectively the system Ranks the identities in the enrolled dataset. The CMC curve illustrates how frequently, on average, the right person ID is included in the best K matches against the training set for each test image, according to \cite{nambiar2019gait}. With the hypothesis that only one sample class in the gallery matches the query, the PRe-ID job is therefore viewed as a recognition issue. As a consequence, a sorted list of gallery classes is provided as the PRe-ID output, depending on some matching resemblance to the query probe.

\subsection{Testing protocol and  settings}

PRe-ID the person retrieval task means finding a person interested in a query image from an extensive dataset.To achieve this goal and to evaluate the proposed approach, three scenarios are used as follow:

\begin{enumerate}
	\item \textbf{Scenario one :} Employs CNN features based AlexNet \cite{han2017pre} using TXQDA. From the CNN fully connected layer 7 (FC7),  4096-dimensional feature vector is extracted for each pedestrian image, and we subdivided it into four vectors (feature parts) of 1024-dimensional to get our tensor with three modes, where: 
	\begin{itemize}
		\item Mode one : consists of a single feature vector with a size of 1024.
		\item Mode two : consists of four vector represent the CNN feature parts. 
		\item Mode three : related to the number of images in each dataset.
	\end{itemize}
	
	
	\item \textbf{Scenario two :} LOMO features are considered with TXQDA.
	The feature vector with 26960-dimensional is subdivided into four vectors of size 6740 to build our $\mathrm{3^{rd}}$ order tensor data with three modes, where:
	
	\begin{itemize}
		\item Mode one: consists of a single feature vector with a size of 6740.
		\item Mode two: consists of four vector represent the LOMO feature parts.
		\item Mode three: related to the number of images in each dataset.
	\end{itemize}
	
	\item \textbf{Scenario three :} CNN+LOMO is considered, the fusion of LOMO and CNN tensors are performed using HDFF to get a new tensor data with three modes: 
	
	\begin{itemize}
		\item Mode one: consists of a single vector feature with a size of 7764 (6740+1024).
		\item Mode one: consists of four vectors representing the feature parts.
		\item Mode three: related to the number of images in each dataset.
	\end{itemize}
	
\end{enumerate}

\subsection{Discussion}

\begin{figure*}[h!]
	\centering
	\setkeys{Gin}{width=.32\textwidth}
	\subfloat[CMC of best features on VIPeR.]{\label{fig:fig5a}\includegraphics{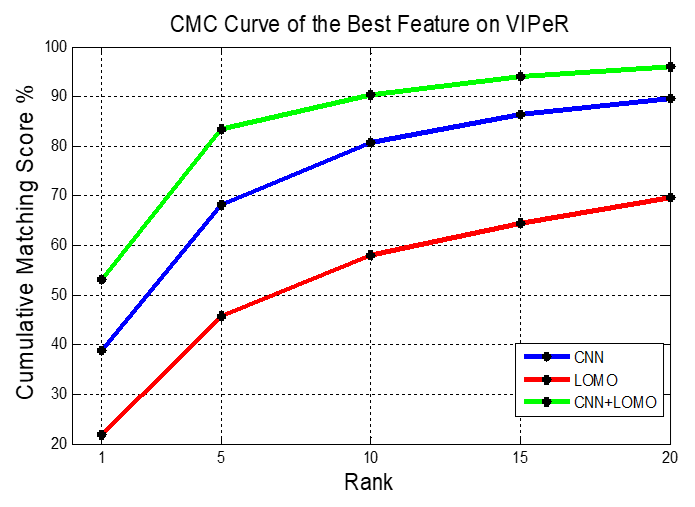}}~
	\subfloat[CMC of best features on GRID.]{\label{fig:fig5b}\includegraphics{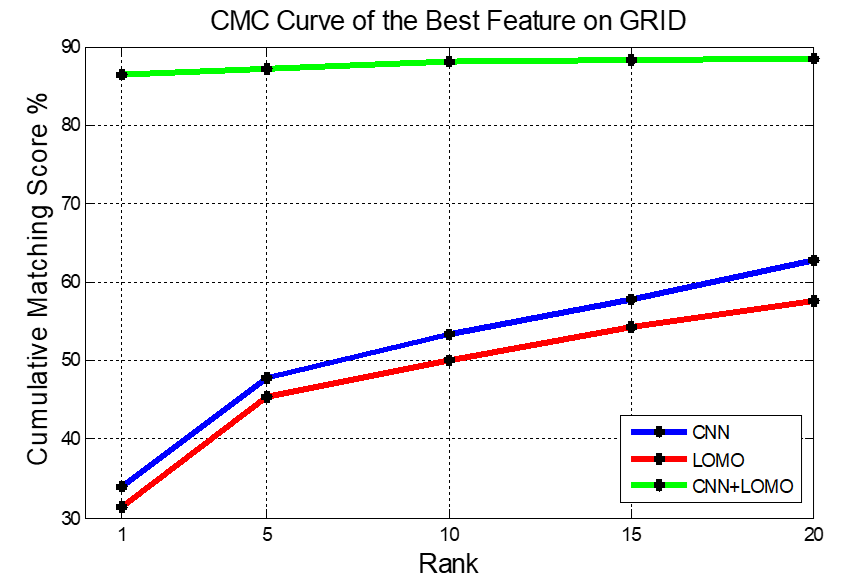}}~
	\subfloat[CMC of best features on PRID450S.]{\label{fig:fig5c}\includegraphics{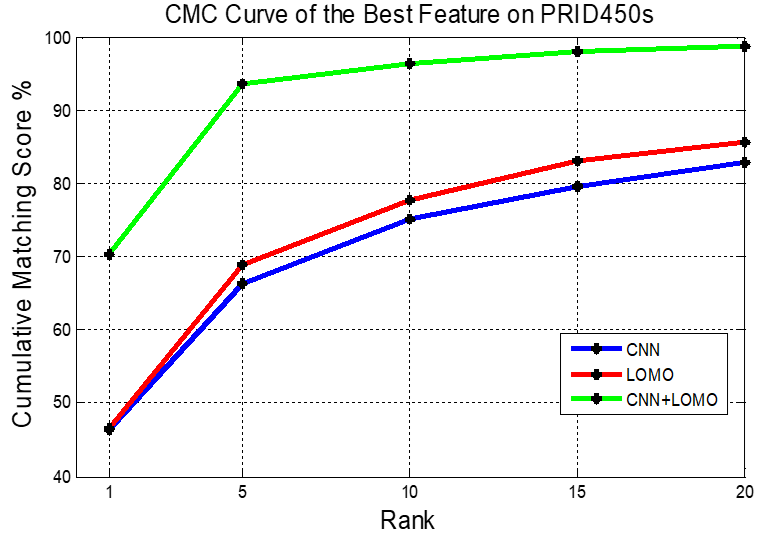}}
	\caption{ CMC scores for Ranks 1 to 20 and different datasets.}
	\label{fig:results}
\end{figure*}

In order to evaluate the performance of the proposed approach and compare it with other  State-Of-The-Art (SOTA) methods, we report CMC as the evaluation metric, at Rank-1, Rank-5, Rank-10, Rank-15, and Rank-20. The obtained CMC curves of the three scenario are shown in the Figures (\ref{fig:fig5a}), (\ref{fig:fig5b}), and (\ref{fig:fig5c}), for the VIPeR, PRID450S and GRID datasets, respectively. In the following, the discussion of the results according to each dataset are presented.

\begin{table*}[ht]
	\caption{All CMC scores (\%) of used datasets GRID considering with CNN, LOMO and the combination CNN+LOMO features.}
	\label{tab:results}
	\begin{adjustbox}{max width=\textwidth}
		\begin{tabular}{|l|ccccc|ccccc|ccccc|}
			\hline
			~ & \multicolumn{15}{c|}{\large\ VIPeR} \\
			
			~ & \multicolumn{5}{c}{CNN} & \multicolumn{5}{c}{LOMO} & \multicolumn{5}{c|}{CNN + LOMO} \\ \hline
			Dim. & Rank-1 & Rank-5 & Rank-10 & Rank-15 & Rank-20 & Rank-1 & Rank-5 & Rank-10 & Rank-15 & Rank-20 & Rank-1 & Rank-5 & Rank-10 & Rank-15 & Rank-20 \\ \hline
			50  & 35.22 & 67.25 & 79.56 & 85.28 & 88.80 & 16.52 & 40.85 & 55.09 & 62.63 & 68.23  & 43.73 & 76.27 & 87.37 & 92.31 & 94.84  \\
			100  & \textbf{38.77} & \textbf{68.16} & \textbf{80.73} & \textbf{86.46} & \textbf{89.65} & 19.40 & 44.87 & 57.50 & 64.94 & 70.03  & 49.81     & 80.95     & 90.44     & 94.30     & 95.92  \\
			150  & 37.31 & 68.45 & 79.43 & 85.89 & 89.78 & 20.73 & 45.41 & 57.75 & 65.13 & 70.38  & 52.12     & 82.69     & 90.82     & 94.53     & 96.20  \\
			200  & 37.22 & 67.18 & 78.67 & 84.87 & 88.80 & 20.98 & 45.89 & 57.66 & 65.00 & 69.87  & 52.31     & 83.64     & 90.70     & 94.21     & 96.11  \\
			250  & 36.71 & 65.98 & 77.18 & 82.94 & 86.87 & 21.20 & 45.85 & 58.20 & 64.56 & 69.84  & \textbf{53.16}     & \textbf{83.32}     & \textbf{90.19}     & \textbf{94.02}     & \textbf{95.82}  \\
			300  & 35.41 & 63.67 & 75.66 & 81.58 & 86.08 & \textbf{21.80}  & \textbf{45.76} & \textbf{58.07} & \textbf{64.49} & \textbf{69.62}  & 52.66     & 83.48  & 90.19     & 93.61     & 95.66  \\
			350  & 34.81 & 62.41 & 73.99 & 80.25 & 84.30 & 21.65  & 45.32 & 57.97 & 64.34 & 69.72  & 52.41 & 83.35 & 89.87 & 93.51 & 95.63  \\ \hline
			\hline
			~ & \multicolumn{15}{c|}{\large\ GRID} \\
			~ & \multicolumn{5}{c}{CNN} & \multicolumn{5}{c}{LOMO} & \multicolumn{5}{c|}{CNN + LOMO} \\ \hline
			Dim. & Rank-1 & Rank-5 & Rank-10 & Rank-15 & Rank-20 & Rank-1 & Rank-5 & Rank-10 & Rank-15 & Rank-20 & Rank-1 & Rank-5 & Rank-10 & Rank-15 & Rank-20 \\ \hline
			50 & 31.7 & 45.5 & 50.5 & 56.00 & 59.28 & 25.52 & 38.1 & 44.5 & 49.36 & 53.5  & 86.48 & 87.04 & 87.68 & 88.16 & 88.72  \\
			100 & \textbf{33.92} & \textbf{47.84} & \textbf{53.44} & \textbf{57.76} & \textbf{62.72} & 29.28 & 41.68 & 46.64 & 50.72 & 55.20  &  \textbf{86.48} & \textbf{87.20} & \textbf{88.00}     & \textbf{88.32}     & \textbf{88.48}  \\
			150 & 33.44 & 47.68 & 54.00 & 58.56 & 62.96 & 30.16 & 43.12 & 48.72 & 53.52 & 57.28  & 86.32 & 87.12 & 87.84  & 88.32 & 88.72  \\
			200 & 33.68  & 48.16 & 54.16 & 58.80 & 64.16 & 31.28 & 43.60 & 48.72 & 52.96 & 56.88  & 86.32 & 87.20 & 88.00 & 88.56 & 88.80  \\
			250 & 33.12 & 47.84 & 53.60 & 58.72 & 63.52 & 31.28 & 44.24 & 49.76 & 54.32 & 57.44  & 86.24 & 87.12 & 87.92 & 88.48 & 88.88  \\
			300 & 32.4 & 47.60 & 53.60 & 58.70 & 63.6 & \textbf{31.44} & \textbf{45.40} & \textbf{50.00} & \textbf{54.30} & \textbf{57.60}  & 86.32 & 87.20 & 87.84 & 88.56 & 88.96  \\
			350 & 31.76 & 46.96 & 53.84 & 58.56 & 63.04 & 32.00 & 45.20 & 50.08 & 54.80 & 58.16  & 86.32 & 87.12 & 87.76 & 88.48 & 88.88  \\ \hline
			\hline
			~ & \multicolumn{15}{c|}{\large\ PRID450S } \\
			
			~ & \multicolumn{5}{c}{CNN} & \multicolumn{5}{c}{LOMO} & \multicolumn{5}{c|}{CNN + LOMO} \\ \hline
			Dim. & Rank-1 & Rank-5 & Rank-10 & Rank-15 & Rank-20 & Rank-1 & Rank-5 & Rank-10 & Rank-15 & Rank-20 & Rank-1 & Rank-5 & Rank-10 & Rank-15 & Rank-20 \\ \hline
			50  & 42.84 & 67.02 & 77.24 & 82.67 & 86.53 & 42.44 & 67.96 & 76.67 & 82.27 & 86.44  & 62.53 & 90.09 & 95.38 & 97.24 & 98.27  \\
			100  & 43.96 & 65.29 & 74.67 & 79.78 & 83.91 &46.53 & 68.98 & 77.78 & 83.07 & 85.78 & 67.73 & 92.76 & 96.53 & 98.09 & 98.80  \\
			150  & 44.36  & 64.31 & 73.29 & 77.69 & 81.42 &  \textbf{46.62} &  \textbf{68.40} &  \textbf{76.62} &  \textbf{81.11} &  \textbf{84.40}  & 69.82 & 93.38 & 96.40 & 98.04 & 98.80  \\
			200  & 45.16 & 63.96 & 72.84 & 77.20 & 80.44 & 45.60 & 66.67 & 74.84 & 79.51 & 83.24  & \textbf{70.40} & \textbf{93.64} & \textbf{96.44} & \textbf{98.04} & \textbf{98.76}  \\
			250  & 44.44 & 64.71 & 73.33 & 77.64 & 80.71 & 44.27 & 65.38 & 73.42 & 77.82 & 81.56  & 69.78 & 93.02 & 96.22 & 97.78 & 98.67  \\
			300  & 46.09 & 65.82 & 74.22 & 78.22 & 81.91 & 42.49 & 63.60 & 71.24 & 76.04 & 79.73  & 69.16 & 92.80 & 96.27 & 97.73 & 98.76  \\
			350  & \textbf{46.36}  & \textbf{66.31} & \textbf{75.16} & \textbf{79.69} & \textbf{82.84} & 40.84 & 62.04 & 69.51 & 73.73 & 77.69  & 69.47 & 92.71 & 96.22 & 97.82 & 98.62  \\ \hline
		\end{tabular}
	\end{adjustbox}
\end{table*}

The experimental results for the three datasets are given in Table \ref{tab:results}. The feature dimensions (Dim) is varied to get the best Rank-1 accuracy.The best Rank-1 accuracy is highlighted in bold. Dim represent the first mode (features) of the tensor data reduced by TXQDA.  

For the VIPeR datasets,the experimental results are given in Table \ref{tab:results} (top part). We can see that the combination of CNN + LOMO based on HDFF with TXQDA increase the Rank-1 ($\Delta$ accuracy) significantly by 14.39\% and 31.36\% for CNN and LOMO respectively, as well as, gives an improvement in Rank-20 of 6.17\% and 26.20\% for CNN and LOMO, respectively.

Regarding the GRID dataset, experimental results are shown in Table \ref{tab:results} (middle part). CNN and LOMO features performed a Rank-1 of  33.92\%  and 31.44\% respectively. LOMO presents a Rank-20  of 57.68\%, while CNN came better with Rank-20 of 62.72\%. However, the best PRe-ID  performance is seen when CNN and LOMO features are merged based on the propoed HDFF, in which, $\Delta$ accuracy
in Rank-1 is equal to 52.56\% and 55.04\% for CNN and LOMO respectively. Also, the fusioned features CNN+LOMO improve the accuracy in  Rank-20 about 25.72\% and 30.88\% for CNN and LOMO, respectively. GRID dataset is considered one of the most difficult datasets in the literature, Hence, these improvements provided by our approach is a very valuable results.

Experimental results associated to PRID450S datasets are also presented in Table~\ref{tab:results} (bottom part). It is clear that the CNN (with 350 dimensions) and LOMO (with 150 dimensions) descriptors performed nearly identically with Rank-1 accuracy around 46\%. A slight improvement is seen in LOMO in Rank-20 with an accuracy of 84.40\% in comparison with CNN that reached an accuracy of 82.84\% . Moreover,  the combined features CNN+LOMO always it achieves the best results with  Rank-1 of 70.40\% giving a $\Delta$ accuracy of 24.04\% and 23.78\% compared to CNN and LOMO, respectively when using each of them separately.

As we can see in both Table ~\ref{tab:results} and Figure \ref{fig:results}, CMC curves prove that our approach has strong discriminative power, so that it outperformed in all datasets by high accuracy in all Ranks. 


\begin{figure}
	\centering
	\includegraphics[width=.5\textwidth]{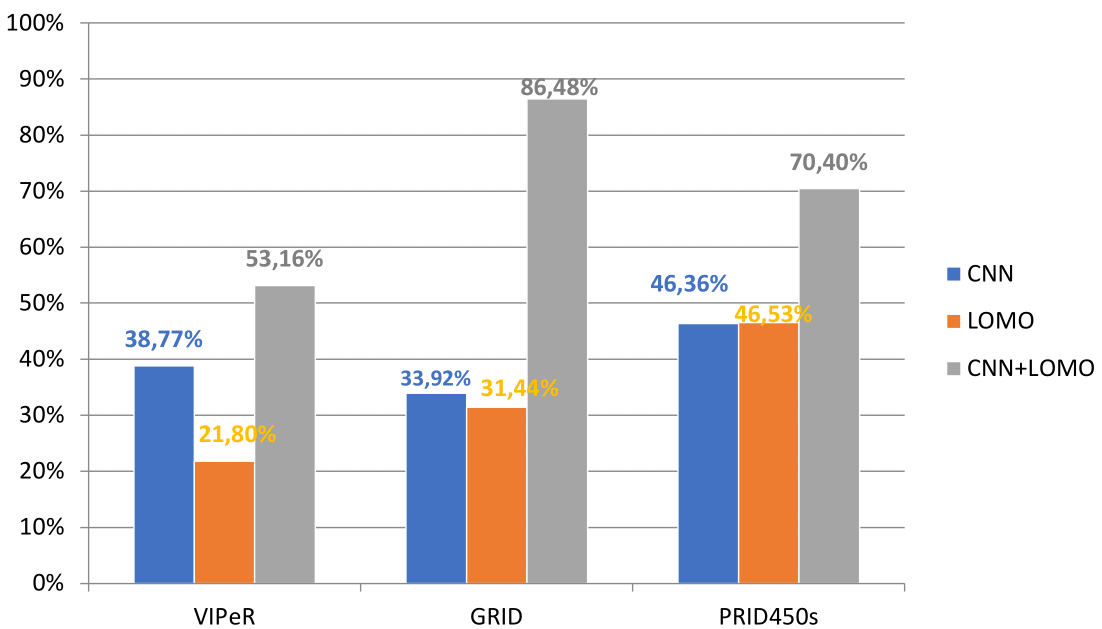}
	\caption{Rank-1 performance on the three datasets VIPeR, GRID and PRID450s.}
	\label{fig:Rank-1}
\end{figure}

\subsection{Impact of datasets on our PRe-ID approach}

In order to thoroughly verify the robustness of suggested method, analyzing the result by changing datasets is crucial. Figure \ref{fig:Rank-1} summarizes all the performance in Rank-1 and compares the obtained results when varying the datasets. The best performance are obtained when applying the proposed approach using CNN+LOMO on all databases, these results highlights the robustness of our method. Notably, with GRID dataset, we have achived a high Rank-1 equal to 86.48\%, which is an impressive result according to its complexity compared to the other two datasets. A Rank-1 of 70.40\% and 53.16\% are obtained with VIPeR and PRID450S, respectively, it is also considered competitive and  excellent results as it will come with us in the comparing of our results with the state-of-the-art.

\subsection{Comparison with the SOTA methods}

The performance of the proposed approach against the SOTA methods via Rank-1 and Rank-20 rates on the three datasets VIPeR, GRID, and PRID450S are summarized in Table \ref{tab:grid}. In this table, the bold results denote the first ranking rates and the second most accurate rates are highlighted in framed values.

For the VIPeR dataset, we have one of the best third results with an accuracy equal to 53.16\% in Rank-1 and  95.82\% in Rank-20. The first best result was achieved by GDNorm ~\cite{2203.01723} with 66.1\% in Rank-1 and 97.4\% achieved by EGOG~\cite{Mortezaie2022} for Rank-20. The second best result is 61.90\% in Rank-1 rate achieved by EGOG ~\cite{Mortezaie2022} and 96.90\% achieved by \cite{7849147} in Rank-20. However, we cannot be certain that the method in EGOG~\cite{Mortezaie2022} is always the best as long as it has not been tested on the GRID database, where most of the methods did not exceed an accuracy of 55.40\%.

The best results for the GRID dataset were obtained using our PRe-ID approach with Rank-1 and Rank-20 accuracies equal to 86.48\% and 88.48\%, respectively. The second best results are for GDNorm ~\cite{2203.01723} with an accuracy equal to 55.40\% in Rank-1. The scheme in~\cite{7849147} achieved the second best result with an accuracy equal to 73.30\%  for Rank-20.

For the PRID450S dataset, the aaproach named EGOG proposed by ~\cite{Mortezaie2022} achieved the best results of 80.8\% and 99.3\%, for Rank-1 and Rank-20, respectively. Our proposed method resulted in the second-best development, with accuracies equal to 70.40\% and 98.76\% for Rank-1 and Rank-20, respectively. 

Our proposed approach is better than the remaining existing SOTA schemes when testing it on GRID dataset, and is competitive for the remaining datasets. 
\begin{table*}[!ht]
	\centering
	\caption{Comparison with the SOTA of Rank-1 and Rank-20 identification rates (\%) on the VIPeR,GRID and PRID450S datasets.}
	\begin{tabular}{|l|c|cc|cc|cc|}
		\hline
		\multirow{2}{*}{Approach}  & \multirow{2}{*}{Year}  & \multicolumn{2}{c|}{VIPeR} & \multicolumn{2}{c|}{GRID} & \multicolumn{2}{c|}{PRID450S} \\
		~ & ~ & Rank-1 & Rank-20  & Rank-1 & Rank-20  & Rank-1 & Rank-20  \\ \hline
		DVR~\cite{10.1007/978-3-319-10593-2_45}   & 2014  & - & -  & - & -  & 28.90 & 82.80  \\
		XQDA~\cite{7298832}   & 2015  & 40.00 & 91.00  & - & -  & - & -  \\
		KEPLER~\cite{7289409}   & 2015  & 42.40 & 90.70  & - & -  & - & -  \\
		FT-CNN+XQDA~\cite{7900000}   & 2016  & 42.50 & 92.00  & 25.20 & 64.60  & 58.20 & 94.30  \\
		SSDAL+XQDA~\cite{10.1007/978-3-319-46475-6_30}   & 2016  & 43.50 & 89.00  & 22.40 & 58.40  & 22.60 & 69.2  \\
		MKSSL+LOMO~\cite{7600432}   & 2017  & 31.20 & 72.80  & - & -  & - & -  \\
		MKSSL+GOG~\cite{7600432}   & 2017  & 40.60 & 85.90  & 24.60 & 64.20  & 61.60 & 96.70  \\
		EP+LOMO~\cite{7896623}   & 2017  & 52.84 & 86.71  & - & -  & - & -  \\
		HIPHOP+LOMO~\cite{7849147}   & 2017  & 54.20 & \circled{96.90}  & 26.00 & \circled{73.30}  & - & -  \\
		DIMN~\cite{8953522}   & 2019  & 51.20 & -  & 29.30 & -  & 39.20 & -  \\
		Kernel X-CRC~\cite{Zhang2021} & 2019  & 51.60 & 95.3  & 26.60 & 69.7  & 68.80 & 98.4  \\
		VS-SSL~\cite{JIA2020107568}   & 2020  & 43.90 & 87.80  & - & -  & 63.30 & 97.00  \\
		DLA~\cite{8953522}   & 2020  & 50.89 & 94.87  & - & -  & - & -  \\
		EML~\cite{Ma2019}   & 2020  & 44.37 & -  & 19.47 & -  & 63.58 & -  \\
		Visual-DAC~\cite{10.1016/j.patcog.2021.108287}   & 2022  & 39.70 & -  & 34.60 & 69.10  & - & -  \\
		STAR-DAC~\cite{10.1016/j.patcog.2021.108287}    & 2022  & 40.80 & -  & 41.20 & 70.60  & - & -  \\
		BNTA~\cite{2203.00672}    & 2022  & 57.40 & -  & 51.10 & -  & - & -  \\
		GDNorm~\cite{2203.01723}   & 2022  & \textbf{66.10} & -  & \circled{55.4} & -  & - & -  \\
		EGOG~\cite{Mortezaie2022}   & 2022  & \circled{61.90} & \textbf{97.4}  & - & -  & \textbf{80.8} & \textbf{99.30}  \\ \hline
		\textbf{CNN + TXQDA (ours)}  & 2022  & 38.77 & 89.65  & 33.92  & 62.72  & 46.36 & 82.84  \\
		\textbf{LOMO + TXQDA (ours)} & 2022  & 21.80 & 69.62  & 31.44 & 57.68  & 46.53 & 85.78  \\
		\textbf{CNN + LOMO + TXQDA (ours)}  & 2022  & 53.16 & 95.82  & \textbf{86.48} & \textbf{88.48} & \circled{70.40} & \circled{98.76} \\ \hline
	\end{tabular}
	\label{tab:grid}
\end{table*}

\section{Conclusion}
\label{sec:conclusion}
In this paper, we proposed a novel strategy for tensor data fusion that strengthens the feature representations of pedestrian images to improve PRe-ID system. By presenting a corresponding multilinear learning approach TXQDA, two efficient features LOMO and CNN are learned to boost the discrimination power. Indeed, our approach allows the integration of several types of features with different dimensions. Numerous experiments confirmed that the proposed framework performs better than the SOTA methods, which highlights that the idea of $\mathrm{3^{rd}}$ order tensor data fusion to improve PRe-ID performances is very interesting. It would be attractive as a future work to apply our approach in vehicle Re-Identification, this important field is characterized by weak features in vehicle images because of the great similarity between the candidates.

\bibliographystyle{elsarticle-num}
\bibliography{references}

\end{document}